\pdfoutput=1

\documentclass[11pt]{article}

\usepackage[]{acl}

\usepackage{times}
\usepackage{latexsym}

\definecolor{OliveGreen}{rgb}{0,0.6,0}

\usepackage{graphicx}
\usepackage{float}
\usepackage{subfig}
\usepackage{booktabs}
\usepackage{amsmath}
\usepackage{CJKutf8}
\usepackage{multirow}
\usepackage{colortbl}
\usepackage[normalem]{ulem}
\useunder{\uline}{\ul}{}

\usepackage[T1]{fontenc}

\usepackage[utf8]{inputenc}

\usepackage{microtype}

\title{Why do objects have many names?\\A study on word informativeness in language use and lexical systems}

\author{Eleonora Gualdoni\footnotemark[1]\\
  Universitat Pompeu Fabra \\
  \texttt{eleonora.gualdoni@upf.edu} \\\And
  Gemma Boleda \\
  Universitat Pompeu Fabra \\
  ICREA \\
  \texttt{gemma.boleda@upf.edu} \\}

\begin{document}
\maketitle

\renewcommand{\thefootnote}{\fnsymbol{footnote}}
\footnotetext[1]{Currently at Apple.}
\renewcommand{\thefootnote}{\arabic{footnote}}

\begin{abstract}
Human lexicons contain many different words that speakers can use to refer to the same object, e.g., \textit{purple} or \textit{magenta} for the same shade of color. On the one hand, studies on language use have explored how speakers adapt their referring expressions to successfully communicate in context, without focusing on properties of the lexical system. On the other hand, studies in language evolution have discussed how competing pressures for informativeness and simplicity shape lexical systems, without tackling in-context communication. We aim at bridging the gap between these traditions, and explore why a soft mapping between referents and words is a good solution for communication, by taking into account both in-context communication and the structure of the lexicon.
We propose a simple measure of informativeness for words and lexical systems, grounded in a visual space, and analyze color naming data for English and Mandarin Chinese.
We conclude that optimal lexical systems are those where multiple words can apply to the same referent, conveying different amounts of information. Such systems allow speakers to maximize communication accuracy and minimize the amount of information they convey when communicating about referents in contexts.
\end{abstract}

\section{Introduction}

A pervasive property of human lexical systems is that many names can be assigned to the same object. In other words, our semantic system allows for a soft mapping between referents and words \citep{Rosch1975, Snodgrass1980, Graf2016, GualdoniJML}. For instance, speakers can call the same chip \textit{purple} or \textit{magenta} \citep{Monroe2017}, and the same animal \textit{dog} or \textit{Dalmatian} \citep{Graf2016,Silberer2020mn}. 

At the same time, a large body of literature has claimed that human lexicons are optimized for efficient communication, which implies allowing for accurate communication exchanges while maintaining a compact size \citep{Regier2015, Xu2020, Zaslavsky2018}. The existence of a soft mapping, which is not the most compact solution possible, may appear on a first glance at odds with such a pressure for efficiency. In this paper, we ask: is a soft mapping between referents and names an efficient solution?
In an analysis of color naming data for English and Mandarin Chinese, we show that, indeed, at least for our domain of interest, a soft mapping is an efficient solution in that it achieves a good a trade-off between the amount of information that speakers have to convey in their \textit{contextual} interactions, on the one hand, and the overall communicative accuracy they can achieve, on the other.

\begin{figure}[!t]
  \centering
\includegraphics[trim={4cm 1cm 8cm 6cm}, clip=true, width=0.75\linewidth]{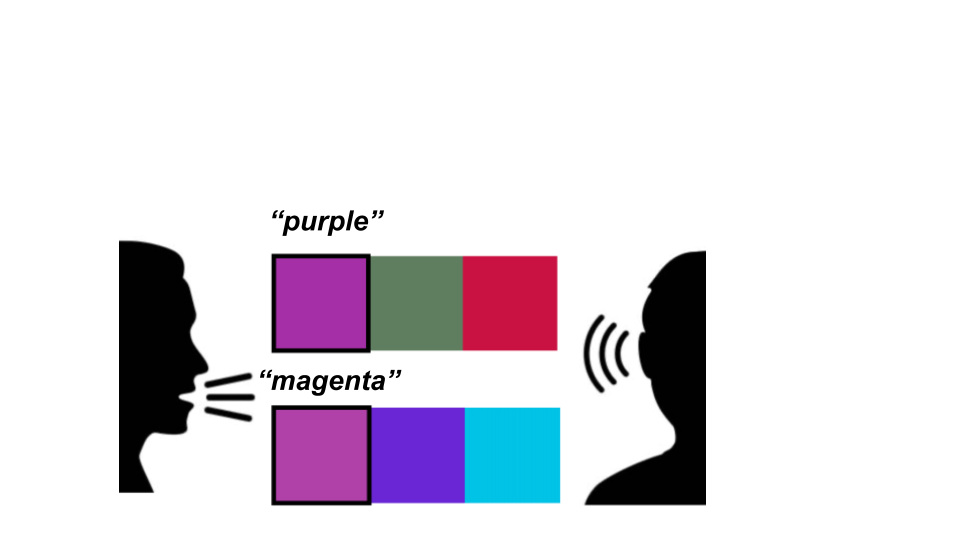}
 \caption{To allow successful identification of a target color chip (in the black frame) within a grid of candidates, a general term like \textit{purple} is sufficient when the context is not challenging (above). A more specific name like \textit{magenta} is needed when the distractors compete more with the target (bottom) ---data from \citet{Monroe2017}.}
  \label{fig:communication}
\end{figure}

Indeed, communication exchanges between interlocutors take place in rich visual contexts.
The dynamic nature of our environment and of speakers' goals constrains naming choices. In situations where it's essential to distinguish one item from context objects, some names can be better than others \cite{Graf2016, Monroe2017, madebach2022}: for instance, when we want our florist to hand us a bouquet of our favourite flowers, the word \textit{flowers} does not provide enough information, while the word \textit{daisies} does.
\citet{Monroe2017} collected experimental data on the phenomenon, which we analyze in the current study. They asked pairs of speakers and listeners to communicate about target color chips appearing in a grid surrounded by distractor chips ---see 
Figure \ref{fig:communication}.
When the target chip is easily distinguishable from the distractors (top), a general term like \textit{purple} might suffice. However, in a more challenging context, where target and distractors are similar (bottom), a more precise term like \textit{magenta} might be necessary to ensure successful communication.
What are the consequences of this on the structure of lexical systems?

Good lexical systems need to be simple, which minimizes cognitive load, and informative, which maximizes communicative effectiveness \cite{Regier2015}. 
Studies have formalized this principle within an information-theoretical framework, showing that human systems optimize a trade-off between the amount of information provided and system complexity \cite{Regier2015, Zaslavsky2018, Xu2020, Maldonado2021}.
While these studies often account for flexible semantic mappings \cite{Zaslavsky2018}, 
they do not study communication as \textit{situated}, with speakers and listeners interacting in an always changing environment.

In this work, we explore why a soft mapping between referents and names is a good solution for in-context communication. 
Note that most research on the semantic properties of the lexicon has so far focused on a different aspect of the soft mapping between language and its use, complementary to the one we study here: ambiguity and polysemy, or the fact that most words have  multiple meanings \cite{juba2011compression,piantadosi+etal:2012,Regier2015,o-connor15:ambiguity-good}.
We reverse the question, asking why a referent can be described with different words \cite{Graf2016}. This phenomenon entails that similar, overlapping meanings can be denoted by different words.

Our method introduces a measure of \textbf{word informativeness} based on word denotations and grounded in a visual space, which can also be used to measure the information provided by lexical systems as a whole. 
With it, we analyze the color naming systems of English and Mandarin Chinese, and claim that their structure is key to achieve successful communication \textit{in context}, with interlocutors communicating in differently challenging situations. 
We first replicate findings from previous studies, showing how speakers adjust their lexical choices to context pressures, leveraging a flexible mapping between referents and words (\textbf{\textit{I} and Language Use}).
We then move to the system level, and show that alternative systems with no such flexible mapping are sub-optimal (\textbf{\textit{I} and Language Systems}).\footnote{Scripts are available at \url{https://osf.io/n3cxh/}.}

\section{Related work}
\label{sec:RW}

Studies modeling language use have explored how speakers adapt their referring expressions and naming choices to the local context in which target referents appear \cite{Graf2016, Monroe2017, Degen2019, madebach2022} or to their communicative goal \cite{vanderwege2009, madebach2022}. These patterns have been formalized in the unified quantitative framework of Rational Speech Act theory \cite[RSA;][]{Frank2012, Goodman2016, Franke2016, Graf2016, Degen2019}.%
\footnote{Another theoretical framework that places emphasis in speaker-hearer interaction mechanisms is Bidirectional Optimality Theory \cite{blutner2003optimality,bi-opt-theory2011}.}

RSA models focus on the \textbf{contextual informativeness} of referring expressions and utterances: the information that a word provides is measured \textit{in context}, factoring in similarities and differences between a target referent and context objects. If a target object, e.g., a dog, appears in a context surrounded by other dogs, the word \textit{dog} will not provide enough information about the target referent, and speakers will avoid it, choosing a more specific expression like \textit{Dalmatian}, in order to help listeners identify the target \cite{Graf2016}. Speaker production choices are also constrained by considerations about utterance cost, often measured in terms of utterance length. 
Speakers are hypothesized to choose referring expressions to maximize a utility function, trading off the maximization of the contextual informativeness with the minimization of the production costs. 
The RSA tradition, given this major focus on context-dependent word informativeness, does not discuss properties of lexical systems as a whole.

 
Cross-linguistic studies on lexical systems have highlighted that, even though different languages partition their semantic space in different ways, this variation is constrained. The structure of lexical systems is believed to derive from the competing communicative principles of informativeness and simplicity, and languages optimize this trade-off ---similar in nature to the one discussed for language use--- in different ways \cite{Regier2015, Zaslavsky2018, Xu2020, Maldonado2021}. In this tradition, rooted in rate-distortion theory, the informativeness of a word is inversely related to the reconstruction error caused in a listener when a speaker uses it to describe a referent. In this sense, word informativeness is \textbf{non-contextual}: the informativeness of a word \textit{w} for a target object \textit{t} relates to the word's semantics and the referent's properties, and is not conditional on the \textit{local context} in which \textit{t} appears. This feature is in common with the word informativeness measure we adopt in this study. 
Of note, \citet{Zaslavsky2020rdrsa} showed theoretical connections between the objective proposed in the RSA framework and rate-distortion theory, suggesting that similar pressures guide the evolution of lexical systems and their pragmatic use (on a similar topic, see also \citealt{Brochhagen2018}). 

In this work, we propose a new measure of word informativeness that allows us to study speakers' adaptation to context in language use as well as the structure of lexical systems as a whole, bridging a gap between approaches focusing on contextual informativeness and approaches focusing on lexical informativeness ---see Section \ref{sec:measure}.


\section{Methods}

\begin{figure*}
  

  
\centering
\subfloat[]{\includegraphics[trim={3cm 3cm 1.5cm 2cm}, clip=true, width=0.49\linewidth]{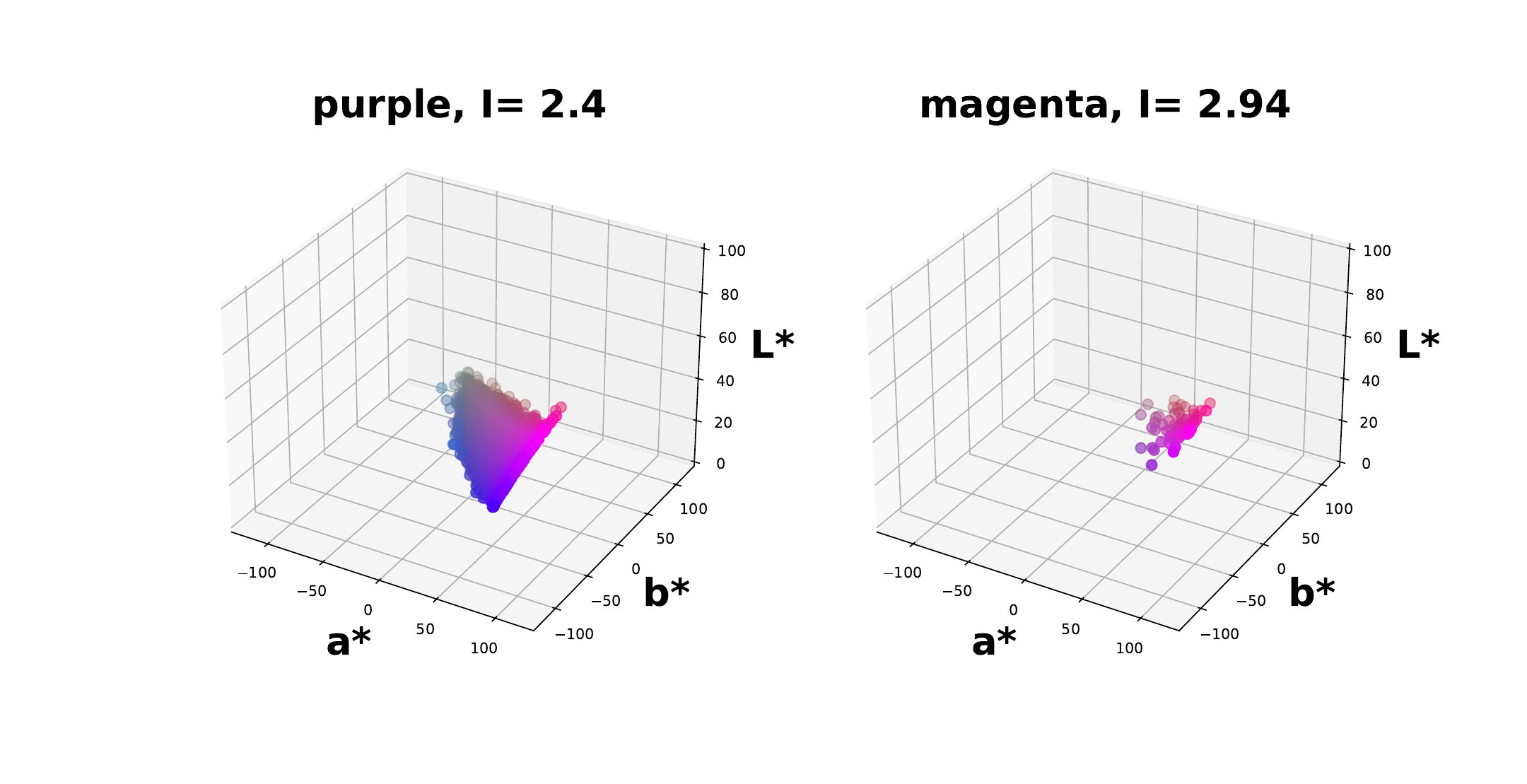}\label{fig:purple_magenta}}
\hfill 
\subfloat[]{\includegraphics[trim={3cm 3cm 1.5cm 1cm}, clip=true, width=0.49\linewidth]{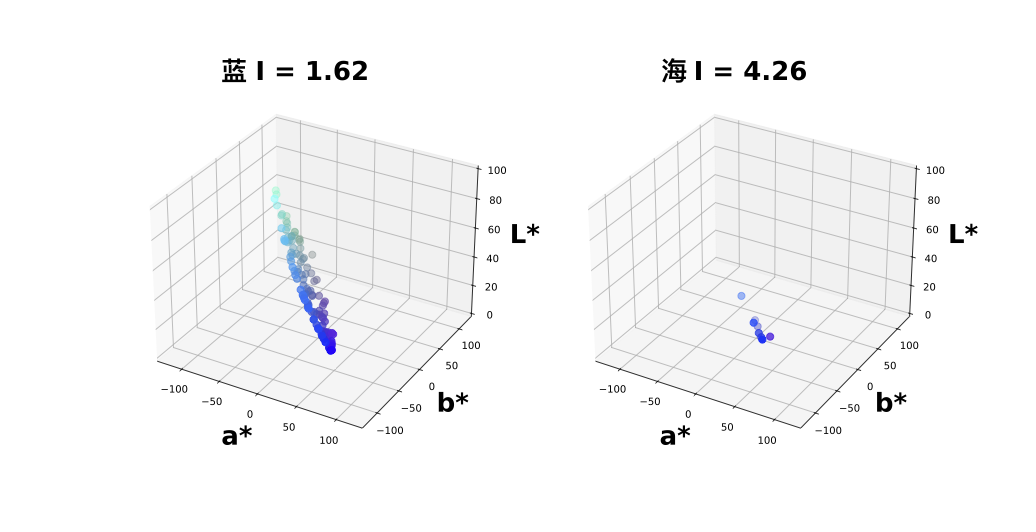}\label{fig:blue_ocean_ch}}

\caption{Denotation in the CIELAB color space of the words \textit{purple} and \textit{magenta} (a) and 
\begin{CJK}{UTF8}{gbsn}
蓝 ``blue” and 海 ``ocean”
\end{CJK}
(b). Note that there is a difference in numbers of objects, that we control for when computing $I$. A color chip called
\begin{CJK}{UTF8}{gbsn}
海 ``ocean”
\end{CJK} (b) would not be located in the top and lighter part of the 
\begin{CJK}{UTF8}{gbsn}
蓝 ``blue”
\end{CJK} denotation region; more specific names denote objects occupying smaller volumes in a visual feature space. Smaller volumes correspond to more information provided by the word to a listener, and higher utterance costs for a speaker. Best viewed in color.}
\label{fig:stimuli} 
\end{figure*}

\subsection{Dataset}

We use \citet{Monroe2017}'s dataset of color chips, including both the English and the Mandarin Chinese data.\footnote{Distributed under a CC-BY 4.0 license.}
The English dataset is annotated with more than $53K$ referring expressions collected in a dyadic reference game. In each round, a target color chip is presented to two players in a grid showcasing two other distractor chips, in random order. One player --\textit{speaker}-- is tasked to unambiguously describe a target chip, allowing the other player --\textit{listener}-- to guess the target chip ---see Figure \ref{fig:communication} for an illustration. 
Crucially, the same color chip is annotated multiple times, in differently hard contexts, as defined by the visual distance between target and distractor chips. Such feature of the data enables the analysis of how speakers adapt their referring expressions to the context. \citet{Monroe2017} found that speakers produce longer and more specific referring expressions in harder contexts (specificity was measured via WordNet; \citeauthor{WordNet} \citeyear{WordNet}).
Since we are interested in analyzing properties of the lexicon, after cleaning the data to remove spelling mistakes and noisy annotations (e.g.\ greetings between annotators), we subset the dataset, considering only rounds that were successfully solved with a single word. This leaves us with 16,168 data points. 
The Chinese dataset, constructed in the same way, is smaller: it contains around $2K$ referring expressions, and $749$ rounds successfully solved with a unique word.

\subsection{Word informativeness}
\label{sec:measure}

We propose a new measure of word informativeness ($I$).%
\footnote{We decided to propose this new measure instead of building on the information-theoretic or the RSA traditions because of its simplicity and adequacy for our research question. Note that integrating both system-level and context-dependent informativeness in the aforementioned frameworks is a challenging problem, which implies defining a multi-objective function that interlocutors are believed to optimize, with a well-defined trade-off between task-general and context-dependent pressures; see \citet{gualdoni_neurips} for a first attempt.}
Our measure is inspired in separate traditions in semantics, which have alternatively highlighted the role of things in the world (denotation) or concepts in the mind.
First, following the emphasis on reference of formal semantics \cite{Dowty1981}, we ground word meaning in the set of objects that the word denotes ---in the case study of this paper, the set of color chips that have been labeled with a given word by the participants.
Second, we assume that meanings are convex regions in a meaning space, as in the more cognitively oriented Conceptual Spaces framework \cite{Gardenfors2001, gardenfors2014geometry}.
This way, we approximate the meaning of a color term in terms of a region in the visual space of colors defined by the specific color chips that have been labeled by the color term by a speaker (\citeauthor{erk2009representing} \citeyear{erk2009representing} is an early example of this kind of approach); and we assume that the region is convex when measuring informativeness, as follows.

The intuition behind $I$ is that smaller volumes in a visual feature space provide more information about a referent than larger volumes: knowing that a referent's visual features are located in a small volume of the space informs a listener about what the referent looks like more than does a large volume. 
In other words, general words, like \textit{purple} in the case of colors, or \textit{person} in the semantic domain of people, are labels for objects that are less similar to each other than the referents of specific words, like \textit{magenta}, or \textit{skier} 
---see for instance the denotation of \textit{purple} \textit{vs} \textit{magenta} in Figure \ref{fig:purple_magenta} and the denotation of 
\begin{CJK}{UTF8}{gbsn}
蓝 and 海
\end{CJK} in Figure \ref{fig:blue_ocean_ch}, and Appendix \ref{app:scatters_colors} for other examples. This results in more specific words being denoted by smaller volumes in a visual feature space \cite{GualdoniJML}, which we posit corresponds to higher amounts of information conveyed to a listener.

Following our conceptualization, we define the informativeness of a word \textit{w} ($I_w$) as follows. Given the denotation of $w$ in the space, we compute a measure of the spread of its visual features ($S_w$), based on the  average distance between pairs of objects $o$ that have been referred to by \textit{w}:
\begin{equation}
\label{eq:spread}
S_w =   \frac{1}{N}\sum_{i}\sum_{j \not= i} d(o_i, o_j)
\end{equation}
Then, we define word informativeness as: 
\begin{equation}
I_w = \frac{1}{S_w}     
\end{equation}

In this work, $d(o_i, o_j)$ is the Euclidean distance in the CIELAB space \citep{CIELAB} between objects $o_i$ and $o_j$ called by $w$. The CIELAB space is a color representation model designed to be more perceptually uniform than other accounts, in which Euclidean distances mirror perceptual distance for the human eye \citep{CIELAB}. $N$ is the number of object pairs.\footnote{Since most $I_w$ scores were of the order of $10^{-2}$, we multiply all the scores by $10^{2}$ for readability.}
The same measure could be applied to other metrics and feature spaces as well, modeling nouns from other domains, or other parts of speech as well, e.g. adjectives.\footnote{This measure is instead not directly not applicable to other parts of speech denoting relations, such as verbs and adverbs, which cannot be easily reduced to regions in a meaning space \citep{gardenfors2014geometry}.} 

In the English portion of \citet{Monroe2017}'s dataset, we compute $I_w$ for each color name $w$ appearing at least $10$ times, and in the Chinese portion we set the threshold at $5$ occurrences.\footnote{Since some color names map to more data points than others, to avoid size effects on the value of $I_w$, we adopt a sampling strategy: if a color name has more than $100$ chips associated, we randomly sample $N$ chips for $T$ times, and average the $I_w$ values obtained for each sample. Our results are robust to different sampling sizes and numbers of iterations. We set $N=100$ and $T=30$.}   
We obtain high $I_w$ scores for words like \textit{olive}, \textit{cyan}, or \textit{lavender}, in English, and 
\begin{CJK}{UTF8}{gbsn}
灰 ``ash” , 橄榄 ``olive” , or 海 ``ocean” 
\end{CJK}
in Chinese; and low $I_w$ scores for words like \textit{blue}, \textit{purple}, and \textit{green}, in English, and
\begin{CJK}{UTF8}{gbsn}
蓝 ``blue” , 橙 ``orange” , or 红 ``red” 
\end{CJK}
in Chinese. Speakers generally prefer to refer to objects at their basic level \cite{Rosch1975, Jolicoeur1984}, such as \textit{blue} or \textit{purple} in the color domain \cite{basic_colors}, making more specific names 
like \textit{magenta} or \textit{turquoise} rare options. Rare words come with higher costs, for instance in terms of reading times \cite{Smith2013} or naming latencies \cite{McRae}. At least in the set of words we study here, words with higher $I_w$, not being basic level categories, are expected to be more costly as well.

To connect language use to lexical systems, we define the informativeness of a lexical system \textit{L} as the average over the $I_w$ of the words uttered to solve \textit{N} interactions:

\begin{equation}
\label{eq:system}
I_L = \frac{1}{N}\sum_{i=1}^{N} I_{w}^i
\end{equation}


\section{$\boldsymbol{I}$ and Language Use}
\label{sec:analysis1}

As discussed in Section \ref{sec:RW}, models of language use predict that, in harder contexts, speakers will utter longer and more specific referring expressions to achieve successful communication. In our analysis of language use, we aim at replicating these findings with our word informativeness measure and \citet{Monroe2017}'s data. Easier contexts are those where targets and distractors share fewer properties \cite{Graf2016, Degen2019} or, in the case of color chips, target and distractor chips are further away in the color space \cite{Monroe2017}. 
Our expectation for $I_w$ is that in harder contexts higher $I_w$ will be needed to reach successful communication ---Figure \ref{fig:communication}, bottom. 
 Recall that we are especially interested in exploring what happens \textit{to the same target} in different contexts. 
 Thus, we subset the data to keep chips that appear at least twice in the dataset (see Appendix \ref{app:all_data} for models fitted on all the data). This leaves us with $5491$ data points across $2524$ target chips, for English, and $60$ data points across $29$ target chips, for Chinese.


\paragraph{Models} 
We use the distance between the target chip and the hardest distractor as a measure of \textbf{context ease}: the larger the distance, the easier the task.\footnote{The distance to the other distractor, in our data, is quite highly correlated with the distance to the closest one, which is supposed to compete more with the target (r$=$0.58, p$<$0.001). Therefore, we only consider the latter.}
We build a linear mixed-effects model, predicting $I_w$ based on context ease. We add random intercepts and random slopes for the target chips (for English and Chinese) and for the worker ids (for English only, since they are not available for the Chinese data).
Our hypothesis, based on previous work, is that easier visual contexts will be characterized by a decrease in $I_w$.

\paragraph{Results}

We replicate findings from previous literature with our $I$ measure and data for English (Table \ref{tab:model_eng_rep}, first row): for the same target, when the context is easier, lower values of $I_w$ allow for communication success. We do not find an effect for Chinese (Table \ref{tab:model_eng_rep}, second row).
Of note, as mentioned above, if we consider only the chips that appear at least twice in the dataset, we only have $29$ possible targets to fit the model on Chinese, which is probably too little to identify an effect (see Appendix \ref{app:all_data} for results on the whole data).

 Considering our results for English, we can see that, on average, for the same chip, an increase of $50$ in context ease leads to a decrease of $0.5$ in name $I_w$. 
 How to interpret this? 
 For $I_w$, this means moving, approximately, from \textit{magenta} to \textit{purple} ($I_w$ $= 2.93$; $I_w$ $= 2.30$) or from \textit{grass} to \textit{green} ($I_w$ $= 3.07$; $I_w$ $= 2.59$).
 As for context ease, in Figure \ref{fig:communication} - top, the distance between the purple target and the green distractor (middle) is 54, while in Figure \ref{fig:communication} - bottom, the distance between the magenta target and the purple distractor (middle) is 17. Therefore, moving from the bottom case to the top case means increasing context ease of 37.%
 \footnote{Here we are considering, for the sake of the example, the chips in the middle as the only distractors.}

\begin{table}
\small
\centering
\begin{tabular}{lccc}
\toprule
 & \textbf{} & \textbf{Estimate} & \textbf{Std. Error} \\
\midrule
\multirow{2}{*}{\textbf{English}} & Intercept & 3.51*** & 0.05  \\
 & Ctx ease & -0.01*** & 0.00 \\
 \midrule
\multirow{2}{*}{\textbf{Chinese}} & Intercept & 2.54*** & 0.36  \\
 & Ctx ease & 0.00 & 0.01 \\
\bottomrule
\end{tabular}
\caption{Fixed effects of the linear mixed-effects model fitted on the English and Chinese data subset of repeated chips. Asterisks express p values: *** = p $< 0.001$. }
\label{tab:model_eng_rep}
\end{table}

\begin{figure}
    \centering
    \includegraphics[width=0.9\columnwidth]{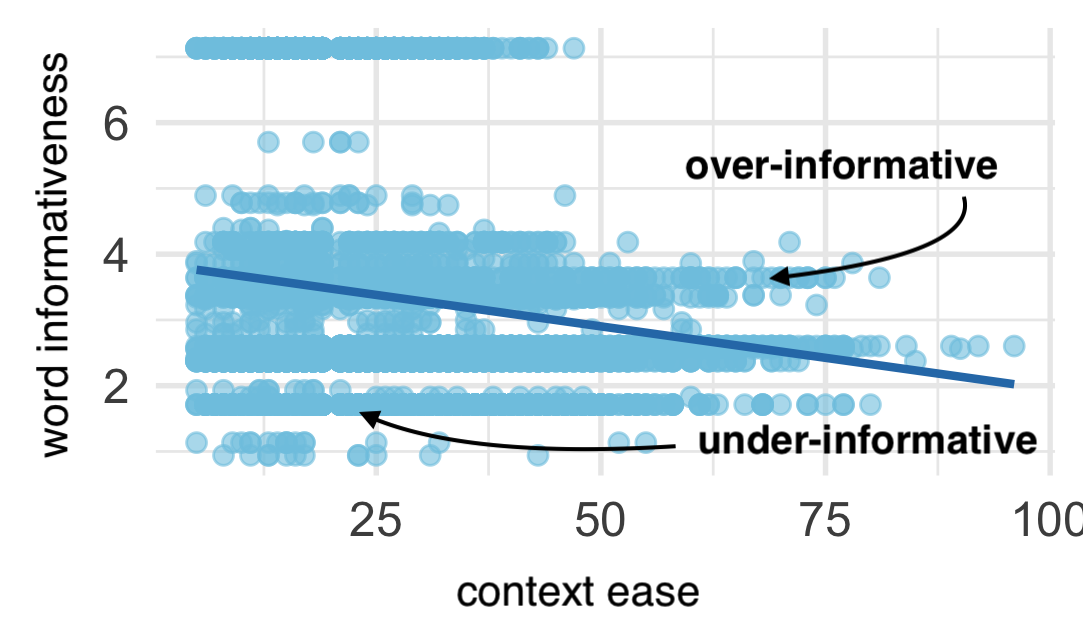}
    \caption{Relationship between context ease and word informativeness ($I_w$) in the portion of \citet{Monroe2017}'s English dataset considered in Table \ref{tab:model_eng_rep}. Communication in easier contexts can be successful with less informative words.}
    \label{fig:scatter}
\end{figure}

The relationship between context ease and $I_w$ for the English data subset is illustrated in Figure \ref{fig:scatter}. Even if the general trend follows our hypothesis, there are some data points that indicate an over-informative (center-right) or under-informative (bottom-left) behavior by speakers, which we analyze next.

\begin{figure}
    \centering
    \includegraphics[width=0.6\columnwidth]{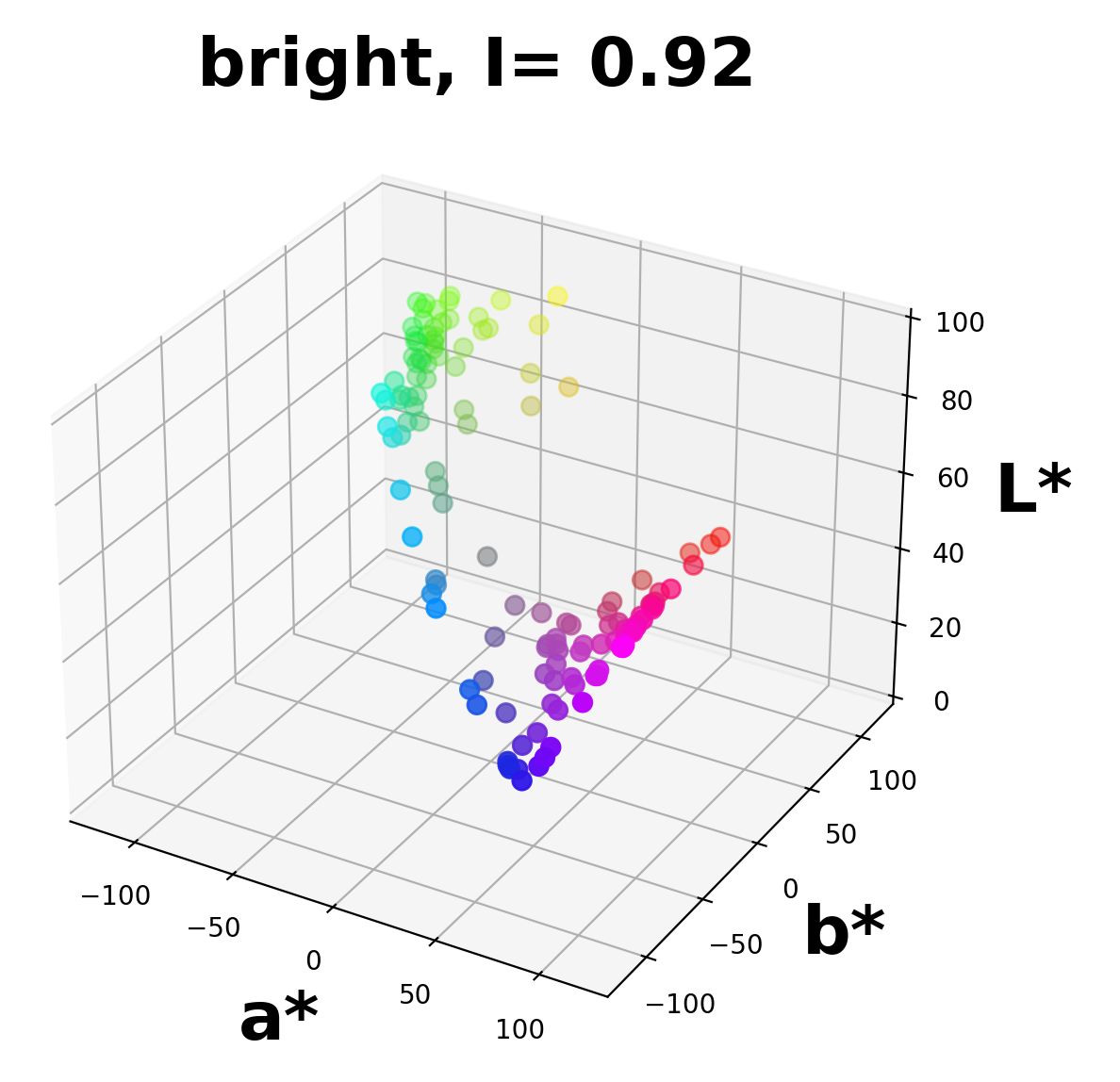}
    \caption{Words like ``bright”  or ``dark” are denoted by non-convex regions, resulting in low informativeness ($I_w$) scores.} 
    \label{fig:bright}
\end{figure}

\paragraph{Analysis of mismatches} 
A qualitative inspection of the mismatches yields two trends. 
First, some interactions in very hard contexts are unintuitively solved with low-informativeness words. The majority of these cases comes from the use of the words \textit{bright}, \textit{dark}, and \textit{light};  or 
\begin{CJK}{UTF8}{gbsn}
亮 ``bright”, 暗 `dark”, or 浅 ``pale”
\end{CJK}. These words are characterized by a non-convex shape in the visual feature space: adjectives like \textit{dark} and \textit{bright} can apply to many different chips that are far from each other in the space ---see Figure \ref{fig:bright}. Our $I$ measure, which is based on the assumption of convexity, results in very low informativeness scores for them.%
\footnote{Of note, given the setup of the referential game these words could be abbreviations for longer and syntactically more complex referring expressions like \textit{dark green} or \textit{the dark one}: different measures may need to be designed to assess the information provided by more complex constructions.}

\begin{figure}[t!]
  \centering
  \subfloat[\textit{mint}: $I_w$ = $3.34$; context ease: $51$]{\includegraphics[trim={0cm 6cm 17cm 5cm}, clip=true, width=0.6\columnwidth]{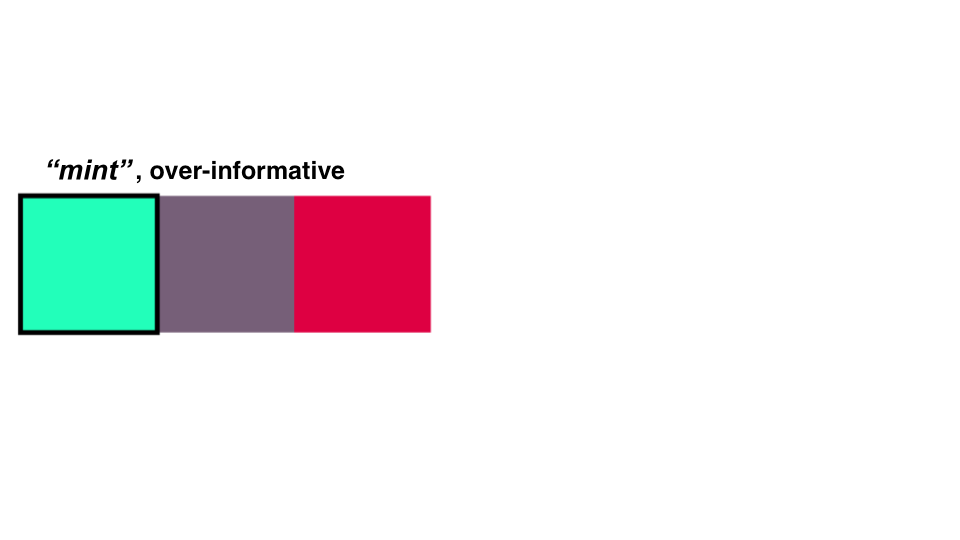}\label{fig:mint}}
  
  \hfill
  
  \subfloat[\textit{blue}: $I_w$ = $1.71$; context ease: $6$]{\includegraphics[trim={0cm 6cm 17cm 5cm}, clip=true, width=0.6\columnwidth]{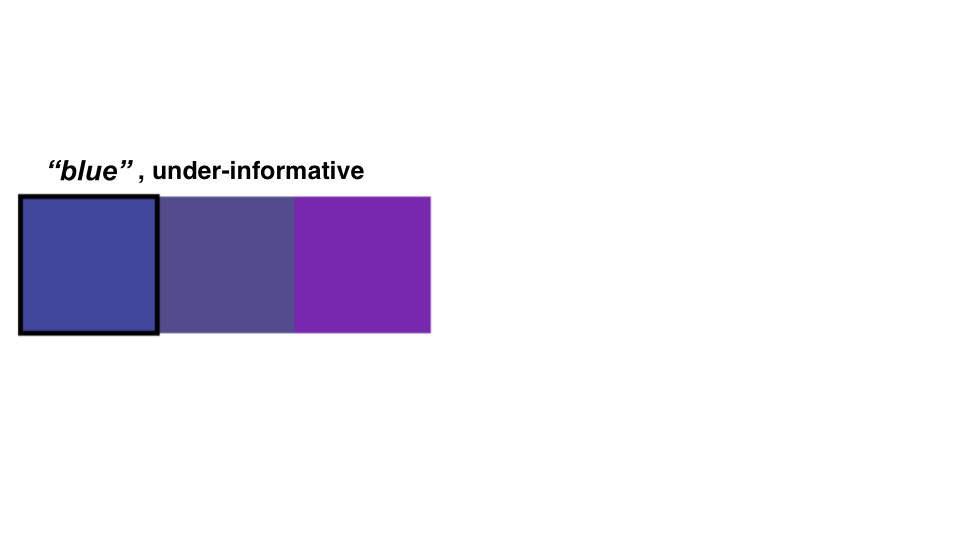}\label{fig:blue}}
  
\caption{Typicality effects in language production. A word with high \textit{I} like \textit{mint} (panel a) can be used when the context is not hard, if the target is very typical for that word. A word with low \textit{I} like \textit{blue} (panel b) can solve the ambiguity in a very hard contexts, if the target is much more typical for the color compared to the distractors.}
\label{fig:typicality} 
\end{figure}

We also find pragmatic effects related to object prototypicality. Psycholinguistic studies have analyzed the effects of prototypicality in descriptive naming tasks \cite{Snodgrass1980, Brodeur2010, Liu2011, tsaparinaRussianNormsName2011, GualdoniJML}, showing that 
the probability of producing a given object name increases with the object's typicality for the name. 
\citet{Graf2016} found this effect also for in-context communication.
They found that speakers deviate from frequent words to use a more costly, specific name when the target is very typical for it, even if a less costly name would suffice to identify the target. In this sense, typicality modulates the cost of the word. We find a similar pattern: when the target is very typical for a name, speakers can be over-informative, producing words with high informativeness, thus more costly, even in cases of easy disambiguation ---see, for instance, the mint chip in Figure \ref{fig:mint}, a context in which ``green" would suffice. 

As for under-informativeness, speakers can produce words with low informativeness in hard contexts, successfully solving the ambiguity anyway ---see, for instance, the blue chip in Figure \ref{fig:blue}. We interpret this as the result of interlocutors' pragmatic iterative reasoning about word interpretations 
\cite{Goodman2016, Graf2016, Degen2019}: in a hard context with multiple chips that could be called \textit{blue}, if the speaker uttered \textit{blue}, it is likely that their intention was to refer to the most prototypical blue. Object prototypicality and interlocutors' reasoning expand the information provided by words beyond denotation, or more generally the semantics of words. In other words, pragmatics enriches word meanings.

\section{$\boldsymbol{I}$ and Language Systems}
\label{sec:analysis2}


We have seen that words with different informativeness values are used by speakers in differently hard contexts. 
What are the consequences of this on the structure of the lexical system? We argue that, to communicate successfully across differently hard interactions, we need a lexical system where multiple entries providing different amounts of information map to the same referent, allowing for a dynamic adaptation in lexical choice. We first formalize this idea in a simulation, and then run an empirical test to confirm it.

\paragraph{Simulation}
Table \ref{tab:real_system} exemplifies a lexical system with a soft mapping between referents and names, listing $6$ referents with $2$ possible names each.
Note that, since general names denote larger volumes in feature spaces, it is more likely for objects to share general names  (e.g.\ \textit{blue}) rather than specific ones (e.g.\ \textit{teal}). However, since lexical systems are complex and not perfectly organized hierarchically, it is also possible to encounter pairs like \textit{referent 2} and \textit{referent 3} that share a specific name (\textit{teal}), but not the general one (\textit{blue} vs.\ \textit{green}). This may be due to prototypicality: \textit{referent 2} may be more typical for the color blue, and \textit{referent 3} for the color green.

\begin{table}[h]
\small
\centering
\begin{tabular}{|c|c|c|}
\hline
\textbf{referent id} & \textbf{general name} & \textbf{specific name}      \\ \hline
referent 1              & blue & turquoise      \\ \hline
referent 2              & blue & teal          \\ \hline
referent 3              & green & teal         \\ \hline
referent 4              & purple & magenta         \\ \hline
referent 5              & purple & mauve         \\ \hline
referent 6              & purple & mauve         \\ \hline
\end{tabular}
\caption{Naming system for 6 hypothetical referents, with a soft mapping between referents and words.}
\label{tab:real_system}
\end{table}

A lexical system like the one just described is not the most compact option: it lists $7$ words for $6$ referents, while, for instance, keeping only the general names would result in $3$ words, and keeping only the specific names would result in $4$ words.
However, as we will show, this kind of system is more efficient: given that referents appear in context, the system can maintain high accuracy in communication, allowing a listener to identify the referent, while minimizing the overall information provided by speakers with their utterances.

\begin{table*}[]
\centering

\begin{tabular}{lcc|cc|cc}
                  & \multicolumn{2}{c|}{\textbf{English - sim}}                                                    & \multicolumn{2}{c|}{\textbf{Chinese - sim}}                               & \multicolumn{2}{c}{\textbf{English - emp.}}                               \\ \cline{2-7} 
                  & \multicolumn{1}{c}{\cellcolor[HTML]{FFFFC7}\textbf{Acc}} & \cellcolor[HTML]{DAE8FC}\textbf{$\boldsymbol{I_L}$} & \cellcolor[HTML]{FFFFC7}\textbf{Acc} & \cellcolor[HTML]{DAE8FC}\textbf{$\boldsymbol{I_L}$} & \cellcolor[HTML]{FFFFC7}\textbf{Acc} & \cellcolor[HTML]{DAE8FC}\textbf{$\boldsymbol{I_L}$} \\ \cline{2-7} 
\textbf{Actual}   & \cellcolor[HTML]{FFFFC7}98\%                              & \cellcolor[HTML]{DAE8FC}2.78       & \cellcolor[HTML]{FFFFC7}99\%         & \cellcolor[HTML]{DAE8FC}1.99       & \cellcolor[HTML]{FFFFC7}96\%         & \cellcolor[HTML]{DAE8FC}3.33       \\
\textbf{General}  & \cellcolor[HTML]{FFFFC7}93\%                              & \cellcolor[HTML]{DAE8FC}2.56       & \cellcolor[HTML]{FFFFC7}93\%         & \cellcolor[HTML]{DAE8FC}1.83       & \cellcolor[HTML]{FFFFC7}81\%         & \cellcolor[HTML]{DAE8FC}2.36       \\
\textbf{Specific} & \cellcolor[HTML]{FFFFC7}96\%                              & \cellcolor[HTML]{DAE8FC}3.99       & \cellcolor[HTML]{FFFFC7}98\%         & \cellcolor[HTML]{DAE8FC}3.13       & \cellcolor[HTML]{FFFFC7}89\%         & \cellcolor[HTML]{DAE8FC}4.24       \\ \cline{2-7} 
\end{tabular}
\caption{Results of accuracy and $I_L$ for actual \textit{vs} hypothetical lexical systems (general words only and specific words only). Column $1$ and $2$: results of simulation; column $3$: empirical data.}
\label{tab:systems}
\end{table*}

How is this achieved? Imagine that each referent can appear in a visual context with another referent, with uniform probabilities (e.g., \{\textit{referent 1}-\textit{referent 2}\}, \{\textit{referent 1}-\textit{referent 3}\}, and so on). Assume furthermore that a listener's accuracy in guessing the target is at chance ($50\%$) if the name uttered to describe the target applies to the distractor as well. Then,
a speaker-listener pair could achieve very high accuracy by leveraging the system structure ---as we have shown in Section \ref{sec:analysis1}--- uttering the general name (low informativeness) when the two referents do not share it (easy context), and the specific name (high informativeness) when they do (i.e.\, in a harder context where the objects share more properties).

We run a simulation with this setup, using the color data of \citet{Monroe2017}.
In particular, we use the target chips that were annotated with at least two different names, and the $I_w$ values of their corresponding names; and we generate from these data all the possible target-distractor pairs. 
Results are reported in Table \ref{tab:systems}, first two columns.

The simulation confirms our hypothesis: for both English and Chinese data, the best accuracy-cost trade-off is achieved by the actual system.
The actual naming system achieves the highest accuracy ($98\%$ English / $99\%$ Chinese) with quite low informativeness $I_L$ ($2.78/1.99$). The only cases where communication success is at chance are those where referents share both general and specific names, akin to the case of \{\textit{referent 5} - \textit{referent 6}\} in Table \ref{tab:real_system}. The other systems are sub-optimal. The hypothetical system keeping only the general name of each referent has a lower $I_L$ ($2.56$ / $1.83$) but achieves a lower accuracy as well ($93\%$ in both cases). 
The hypothetical system keeping only the specific name of each referent achieves an accuracy of $96\%$ / $98\%$, comparable to the one of the actual system (if slightly lower), but exhibits a much higher $I_L$ of $3.99$ / $3.13$.
Mistakes occur in cases where the referents share both specific and general names (as was the case for the actual system), or to the cases where referents share the same specific name, but not the general one (as in the \textit{teal} example above).

Overall, thus, a system with a soft mapping between referents and names is an optimal solution, maximizing communicative accuracy with lower overall $I_L$.

\paragraph{Empirical test}
We collect human data to complement our simulation. We sample $100$ target-distractor1-distractor2 datapoints from the English dataset, uniformly distributed with respect to their context ease, and consider the name that the target received in the sampled triplet as a reference name. To generate lexical systems alternative to the actual one, we simulate for each target a more general name (lower informativeness) and/or a more specific name (higher informativeness). We do so by taking the name with highest or lowest $I_w$ that the same chip received across contexts. This way, we make sure that the word is adequate for the chip.%
\footnote{Given that only a few chips were annotated
more than twice in different contexts with a single
word, in the majority of the cases we either
simulate the general name, or the  specific name.}
In order to measure the accuracies achieved by the different resulting lexical systems, we asked $3$ English native speakers (unrelated to this study) to act as listeners, guessing the target based on the word we provide --see Appendix \ref{app:data_collection} for further details.

Results are reported in Table \ref{tab:systems}, column $3$. The relationship between the listeners' performance in the 3 conditions mirror what we found in our previous simulation.
The actual naming system achieves the highest accuracy ($96\%$), with an intermediate $I_L$ value ($3.33$). This makes it the best system: the general system we simulated comes with lower costs ($I_L = 2.36$), but is not accurate ($81\%$), while the specific system we simulated is more costly ($4.24$), without a gain in accuracy ($89\%$).


Note that the scores in our empirical test are generally lower, which is due to the sampled contexts being harder than in the simulation: 
in the simulation, we created all the possible target-distractor pairs, thus automatically generating a larger number of easier cases, given that each chip has a high similarity only with a few chips, and is visually very different from the majority of the other chips. 
To have a better grasp on our listeners' behavior, we next dive deeper into their mistakes.

\paragraph{Analysis}

\begin{figure}
  \centering
  \includegraphics[trim={0cm, 0cm, 0cm, 0cm}, clip=true, width=0.55\linewidth]{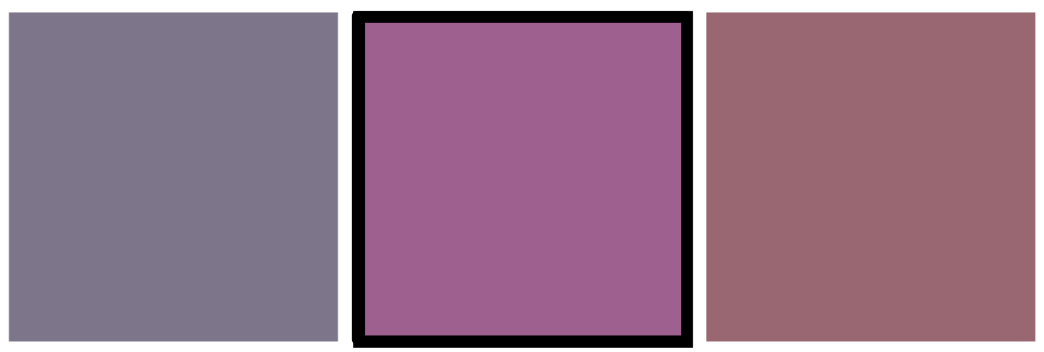}

\caption{Target chip, called \textit{pink} in this context, for which we simulated \textit{mauve}, here misleading, as the rightmost chip was chosen instead.}
\label{fig:simulated_spec} 
\end{figure}

The mistakes that the listeners made are in line with those that arose in the simulation. Figure \ref{fig:simulated_spec} shows an example. The target chip in the black frame, called \textit{pink} in the shown context, was assigned the name \textit{mauve} as a simulated specific name. However, the specific name \textit{mauve} can apply to the rightmost distractor as well, leading to a case where two referents appearing in the same context share the same specific name (as in the \textit{teal} example), and in this case the listener failed to identify the target.%
\footnote{Recall that for some chips we could not simulate a name more specific / general than the actual one (already specific / general). As a sanity check, we report that our annotator's accuracy on this portion of data is in line with that of the actual system ($95\%$ for simulated specific data; $98\%$ for the simulated general data), while it decreases for the portion of data with simulated names ($81\%$ for simulated specific; $69\%$ for the simulated general).}

Cases of this nature, which can result in mistakes in the annotation, are actually good examples of how pragmatics is again at play, expanding word meanings beyond denotational semantics. For instance, both \textit{pink} and \textit{mauve} could describe the target in Figure \ref{fig:simulated_spec} in isolation. The word \textit{mauve} is \textit{per se} more informative than the word \textit{pink}, but it is not \textit{contextually} informative, since the distractor on the right may also be called \textit{mauve}. Given that the target is more prototypical for \textit{pink} than the distractor, a listener may expect the word \textit{pink} --and not \textit{mauve}-- to be used to describe it, even if the context is hard and the word \textit{pink} is less specific. Moving from \textit{pink} to \textit{mauve} increases word informativeness but does not factor in pragmatics, which in this case leads to unsuccessful communication (note that Figure \ref{fig:blue}, discussed in Section \ref{sec:analysis1}, constitutes a successful case of the same type of pragmatic reasoning).

\section{Discussion}

In this work, we have studied why a lexical system where multiple names map to the same referent is a good solution for human communication. We have done so by proposing a measure of word informativeness grounded in a visual feature space and based on word denotations.

Previous studies on the optimality of lexical systems often consider the number of lexical entries in a system as a measure of system complexity \cite{Regier2015, Xu2020}, with smaller lexicons preferable over large ones due to cognitive constraints. These approaches would fail at capturing how an increased lexicon size can become advantageous when we factor in communication in context, allowing for the minimization of the overall amount of information transmitted in language use. Our study, drawing a connection between language use in context and the consequent structure of an efficient lexical system, bridges this gap.

Connecting properties of language production in context with properties of the lexicon is much in the spirit of previous work connecting language production and properties of grammars regarding language universals \cite{hawkins04, franzon-zanini23}.
Future work should explore parallelisms between the lexicon and the grammar in this respect.
This is also intimately connected to diachronic dynamics, and the causes and consequences of semantic change. Adopting an evolutionary perspective, \citet{gualdoni_neurips} study how a human-like semantics can emerge from contextually-rich, pragmatic interactions; and \citet{kobrock-etal-2024-context} compare lexicons emerging in artificial agents that have access to context information to those of context-agnostic ones.
Future work should delve deeper into how word informativeness and reference to objects in context interact to produce a given lexicon. A related question is how system learnability affects communication accuracy and shapes language evolution \cite[on the topic see, for instance,][]{Carlsson2024, Gyevnar2022, tucker2022}.

Our work also resonates with previous research on why the lexicon presents pervasive ambiguity, and specifically the fact that most words have multiple meanings \cite{juba2011compression,o-connor15:ambiguity-good,fortuny-corominas15,piantadosi+etal:2012}.
As mentioned in the Introduction, lexical ambiguity corresponds to one-to-many relationships between words and meanings, while we have focused on many-to-one relationships between words and referents. 
We believe that both phenomena are different consequences of efficiency constraints acting on the lexicon.
Our findings complement previous research \citep{piantadosi+etal:2012} by showing that, in general, many-to-many mappings between words and meanings can be characterized as efficient solutions.

An advantage of our specific approach lies in the simplicity and the flexibility of our informativeness measure, that could be adapted to other kinds of distributed representations, allowing us to study different phenomena besides referential language in language and vision. 
For instance, when the subject of a sentence is unexpected for a listener, a more informative word like \textit{president} instead of \textit{man} may be preferred \citep{aina-etal-2021-referent}. Our measure (which could be derived from language models considering the distances between contextualized embeddings of the same word) could allow for joint analyses of discourse and vision data under the same light, taking a step towards a unified view of human referential acts \citep[see, for instance,][for the analysis of a related phenomenon in morphology]{franzon-zanini23}.

That being said, our measure only accounts for the semantic meaning of words, suffering from a limitation: in every interaction, pragmatics enriches word meanings and modulates the information provided by words in context, or their cost for speakers. A denotational measure of word informativeness cannot capture the full range of phenomena characterizing language production in context. 
Moreover, our formulation of the measure cannot accurately describe words denoted by non-convex regions in meaning spaces. 
We leave it to future work to define alternative measures that take into account more complex shapes (see Figure~\ref{fig:bright}), while still being general enough.

Finally, in our analyses we have made the simplified assumption that speakers want to provide the right amount of information, avoiding over-informative utterances. There is literature showing that this is not always the case \cite{Engelhardt2006, Koolen2011}, and that the production of redundant and over-informative referring expressions can fall in the set of behaviors that maximize efficient communication \cite{Rubio-Fernandez2016, Degen2019}. Moreover, considerations beyond informativeness affect speakers' naming choices, e.g.\ speakers could choose \textit{professor} instead of \textit{woman} to highlight aspects of the referent contingently relevant to them \cite{Silberer2020mn}, even without any explicit pressure for discrimination. A more comprehensive analysis of language use should account for these factors as well. 


\section{Conclusion}
Objects have many names. In this work, we have analyzed human color naming data exploring why some degree of soft mapping is a feature of an optimal lexical system, bridging the gap between analyses of lexical systems and language use. We conclude that systems where multiple words conveying different amounts of information can be used to describe the same referent are optimal, in that they maximize communicative accuracy while minimizing the amount of information conveyed.

\section{Limitations}

Our study aims at analyzing the structure of human lexical systems. However, given the scarce availability of cross-linguistic datasets studying situated communication, we limit ourselves to English and Mandarin Chinese; the latter, with less coverage than the former. This constitutes a limitation of our work, which would benefit from the analysis of a more diverse set of lexical systems.
Along the same lines, our analysis is limited to the color semantic domain: validating the word informativeness measure on a different system of categories, as well as running the analyses on richer semantic domains, would strengthen our conclusions (see \citeauthor{GualdoniJML} \citeyear{GualdoniJML} for preliminary evidence in the domain of people). 

It is also worth noticing that our analysis of the sub-optimality of hypothetical lexical systems is limited to two alternative systems that we could simulate with the data available to us. The overall considerations on the optimality of human-like lexical systems would benefit from the analysis of more, and more diverse, hypothetical alternatives. 

Finally, our simulation in Section \ref{sec:analysis2} relies on simplified assumptions, such as referents co-occurring in contexts with uniform probabilities. Estimating the real probabilities of referents co-occurring constitutes a big challenge, but would also constitute a great improvement for all studies interested in understanding the pressures that shape the human lexical system.

\section*{Acknowledgements}

This research is partially an output of grant PID2020-112602GB-I00/MICIN/AEI/10.13039/501100011033, funded by the Ministerio de Ciencia e Innovación and the Agencia Estatal de Investigación (Spain) and has received funding from the European Union's Horizon 2020 research and innovation programme (grant agreement No.\ 715154).
We thank Thomas Brochhagen, Francesca Franzon and Louise McNally for feedback on an earlier version of the paper.

\bibliography{anthology,custom}
\bibliographystyle{acl_natbib}

\appendix

\section{Details on the data collection}
\label{app:data_collection}

Each annotator was presented with the same set of target chips to annotate, but with different names: one annotator received the actual system, one received the simulated general system, and one received the simulated specific system. Each block of questions contained 5 randomly placed controls, designed to ensure that annotators were paying attention to the task. These cases were made intentionally very simple.
The data collection routine was written in Psychopy \citep{Peirce2019}. There was no time limit for completing the study. 
Instructions for annotators: ``Welcome! 
In this study, we ask you to identify a target color chip in a set of 3 chips, based on a word. You will always see 3 color chips. Above them, there will be a word describing the target. 
We ask you to click on the target. 
Sometimes you will not be sure about your answer. 
Please make your best guess.
Reply with what you think is the most plausible answer”.



\section{Models fitted on all the data}
\label{app:all_data}

Table~\ref{tab:model_eng_all} shows effects when modeling the whole data. For both English and Mandarin Chinese, we identify the expected trend (recall from Section \ref{sec:analysis1}, this effect disappears when we subset the data to keep only chips annotated at least twice across different contexts, which reduces the total of the target chips available to fit the model to $29$).

\begin{table}[H]
\small
\centering
\begin{tabular}{lccc}
\toprule
 & \textbf{} & \textbf{Estimate} & \textbf{Std. Error} \\
\midrule
\multirow{2}{*}{\textbf{English}} & Intercept & 3.80*** & 0.04  \\
 & Ctx ease & -0.02*** & 0.00 \\
 \midrule
\multirow{2}{*}{\textbf{Chinese}} & Intercept & 3.27*** & 0.14  \\
 & Ctx ease & -0.01* & 0.00 \\
\bottomrule
\end{tabular}
\caption{Fixed effects of the linear mixed-effects model fitted on the English data (random intercepts and random slopes for worker-ids) and of the linear model fitted on the Chinese data ---without subsetting the data to include only chips annotated at least twice. Asterisks express p values: *** = p $< 0.001$; * = p $< 0.05$.}
\label{tab:model_eng_all}
\end{table}

\section{Color denotation in visual space}
\label{app:scatters_colors}

\begin{figure*}
  \centering
  \subfloat[]{\includegraphics[trim={3cm 3cm 1.5cm 2cm}, clip=true, width=0.7\linewidth]{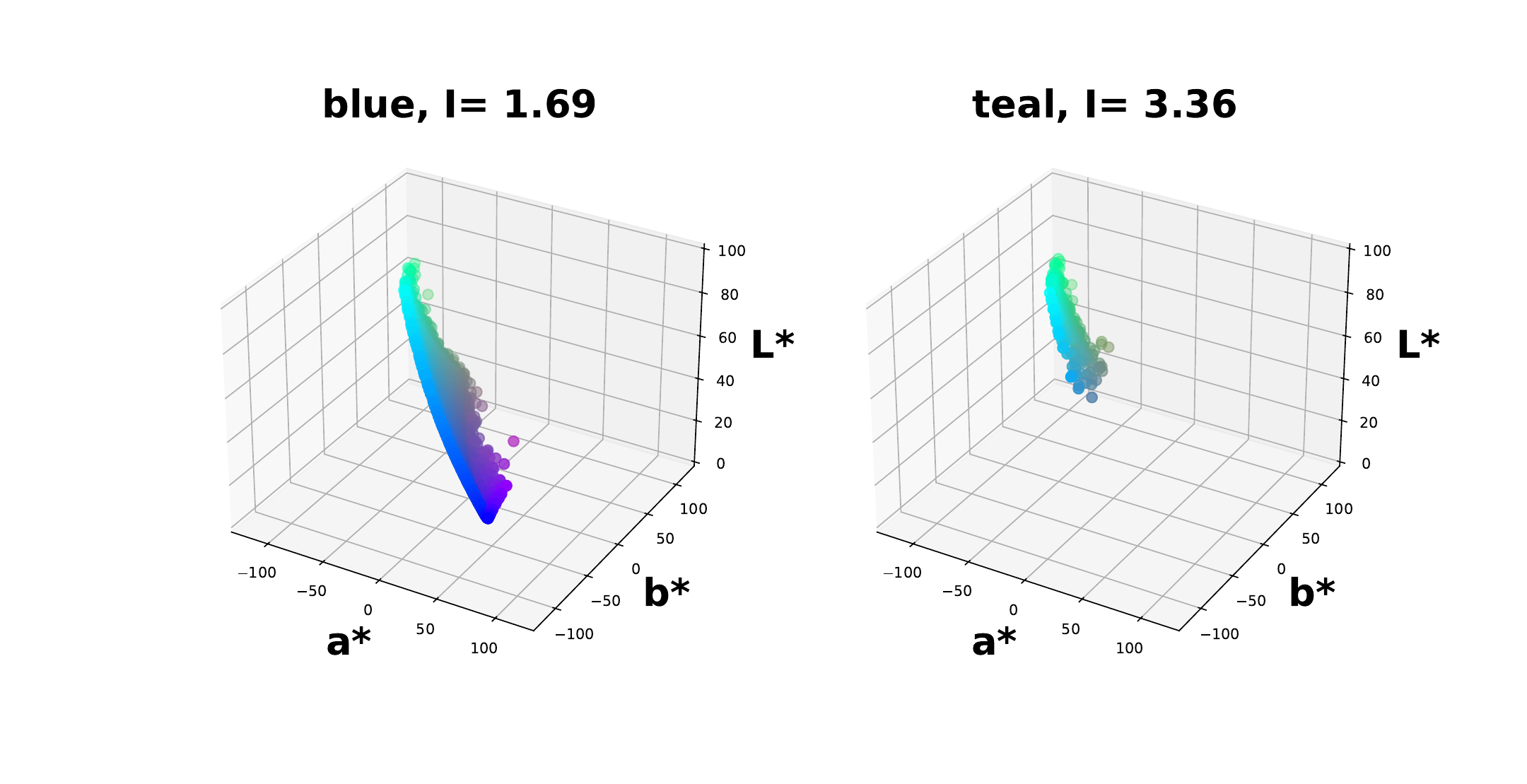}\label{fig:blue_teal}}
  
  \hfill
  
  \subfloat[]{\includegraphics[trim={3cm 3cm 1.5cm 2cm}, clip=true, width=0.7\linewidth]{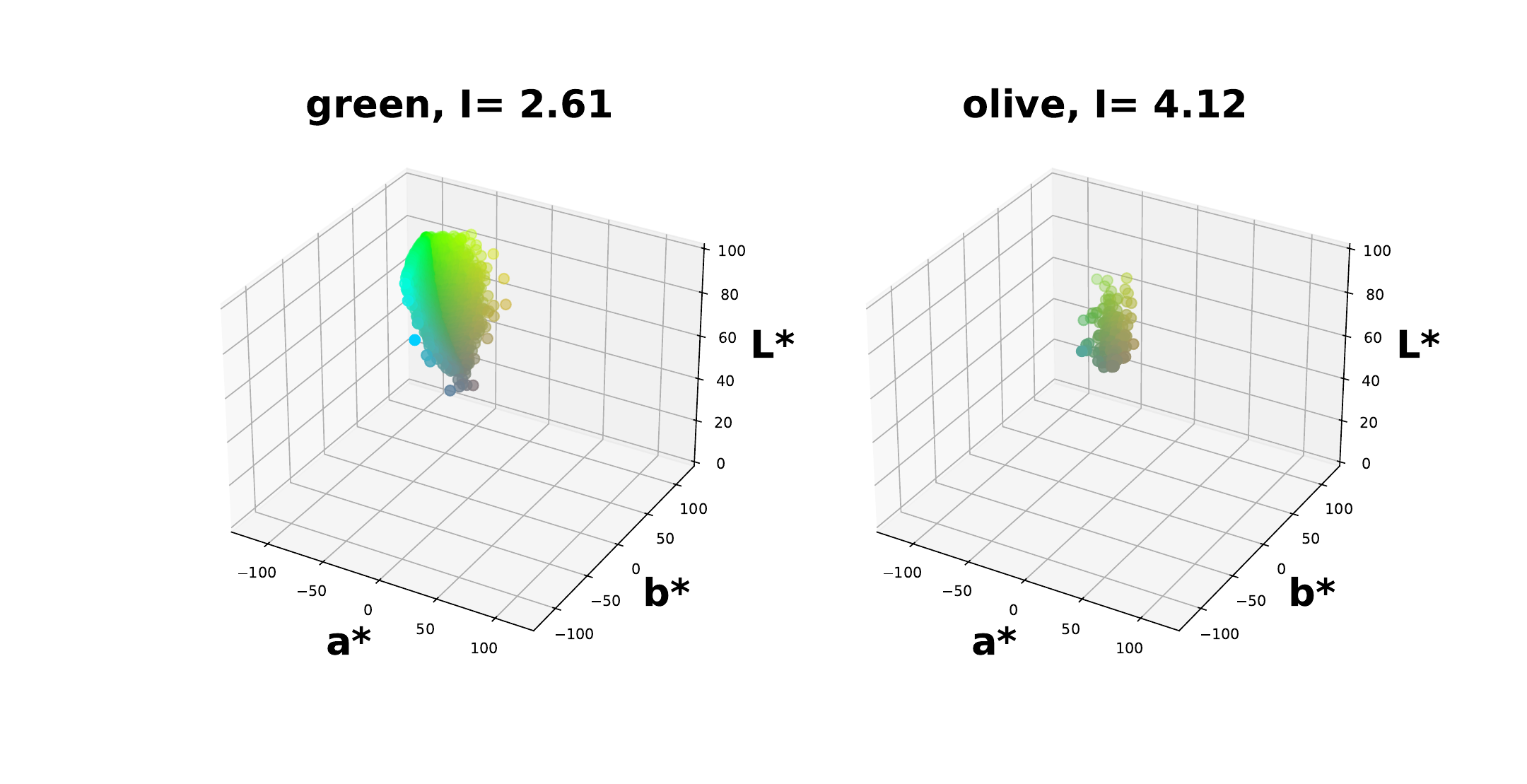}\label{fig:green_olive}}
  
  \hfill

  \subfloat[]{\includegraphics[trim={3cm 3cm 1.5cm 2cm}, clip=true, width=0.7\linewidth]{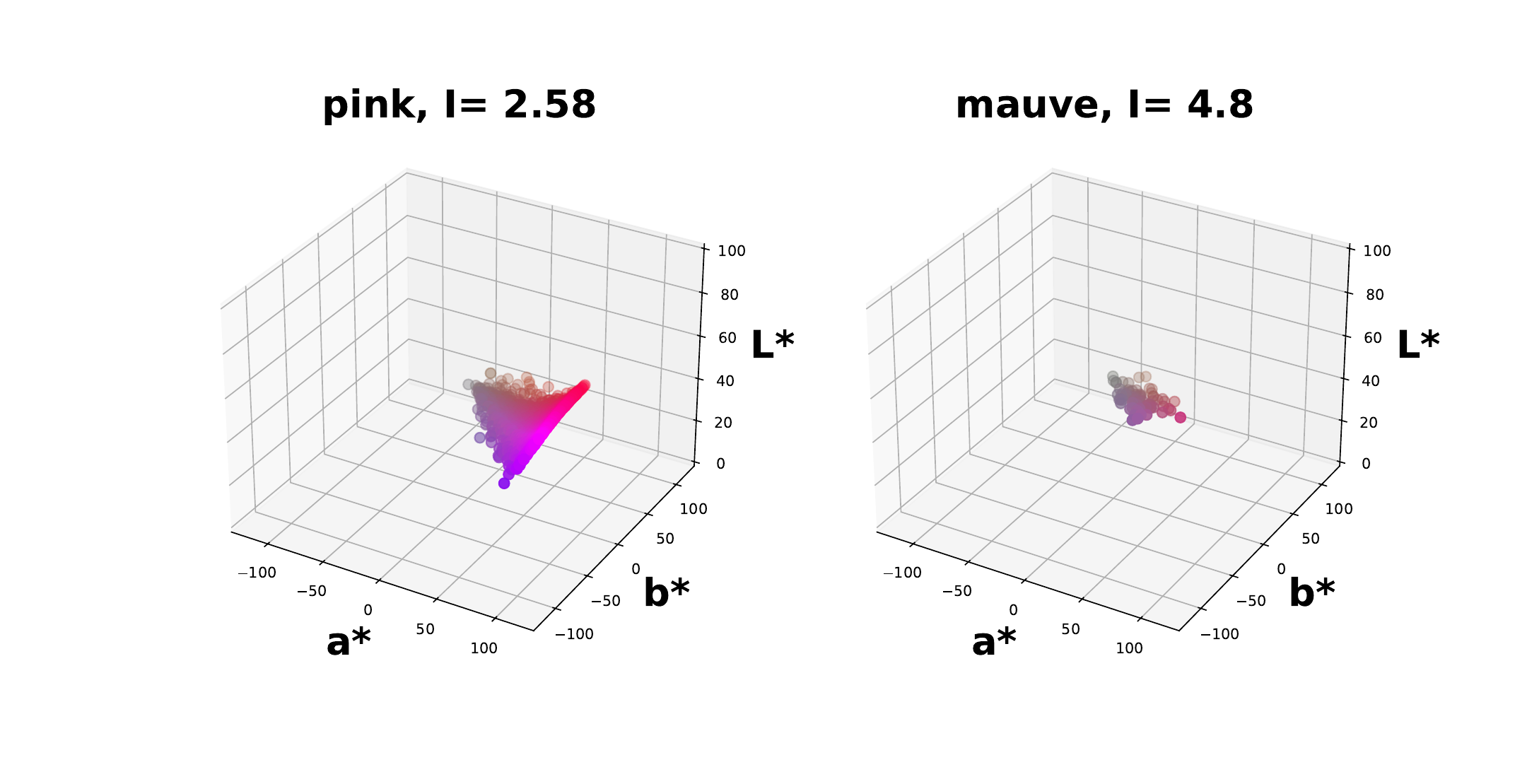} \label{fig:pink_mauve}}
  
\caption{Denotation in the CIELAB color space for \textit{blue} and \textit{turquoise} (panel a), \textit{green} and \textit{olive} (panel b), \textit{pink} and \textit{blood} (panel c).}
\label{fig:colors_English} 
\end{figure*}

\begin{figure*}
  \centering
  \subfloat[]{\includegraphics[trim={3cm 3cm 1.5cm 2cm}, clip=true, width=0.7\linewidth]{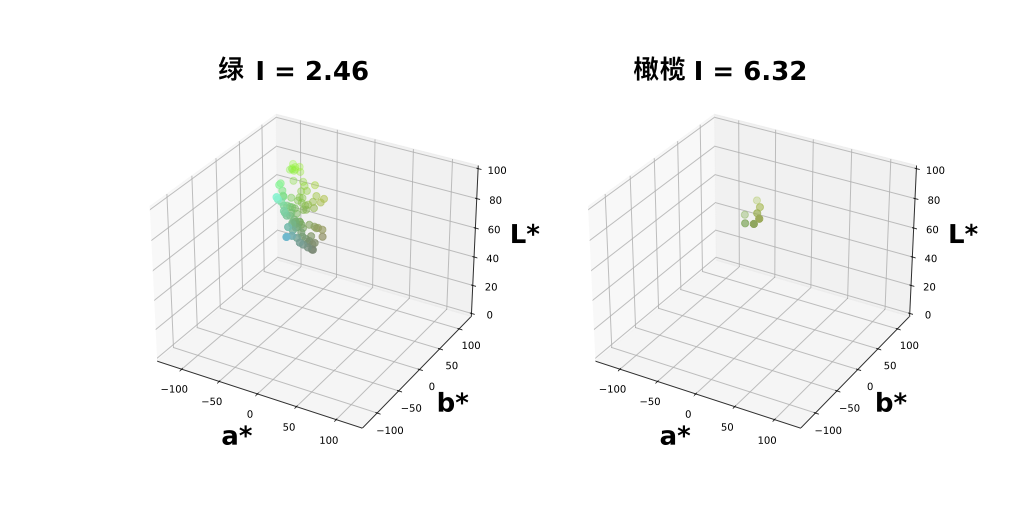}\label{fig:green_olive_Ch}}
  
  \hfill
  
  \subfloat[]{\includegraphics[trim={3cm 3cm 1.5cm 2cm}, clip=true, width=0.7\linewidth]{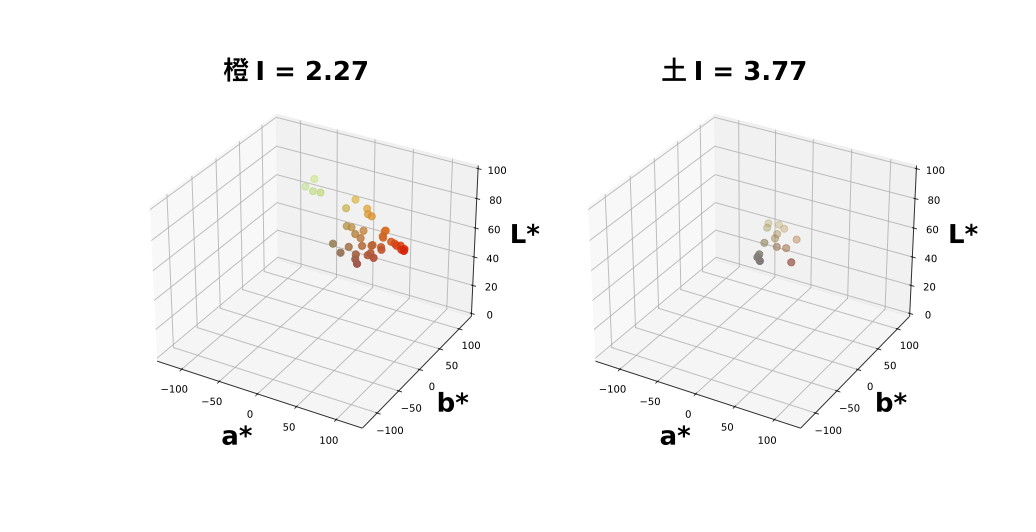}\label{fig:orange_soil_Ch}}
  
  \hfill

  \subfloat[]{\includegraphics[trim={3cm 3cm 1.5cm 2cm}, clip=true, width=0.7\linewidth]{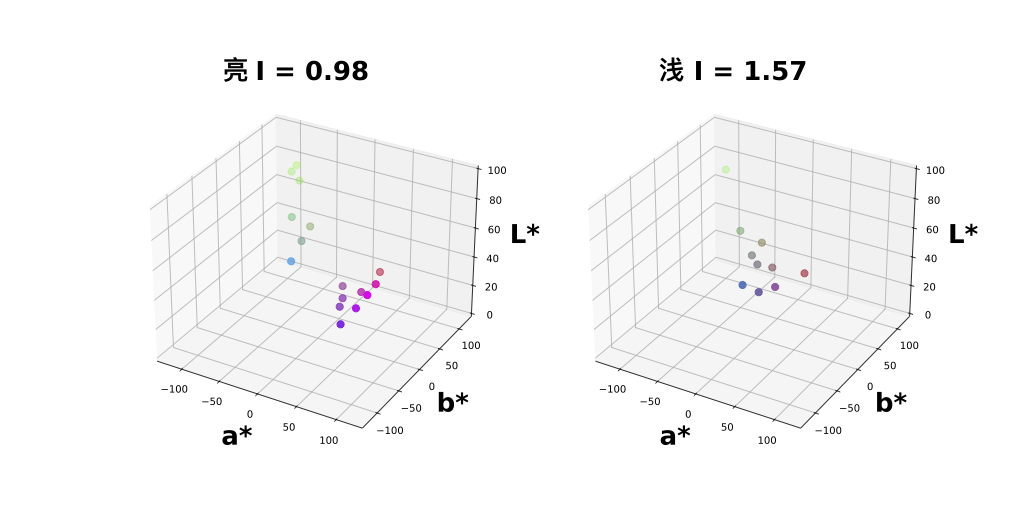} \label{fig:bright_pale_Ch}}
  
\caption{Denotation in the CIELAB color space for 
\begin{CJK}{UTF8}{gbsn}
绿
\end{CJK}
and 
\begin{CJK}{UTF8}{gbsn}
橄榄
\end{CJK} (panel a),   
\begin{CJK}{UTF8}{gbsn}
橙 
\end{CJK} and
\begin{CJK}{UTF8}{gbsn}
土 
\end{CJK} (panel b), 
\begin{CJK}{UTF8}{gbsn}
亮 
\end{CJK} and
\begin{CJK}{UTF8}{gbsn}
浅 
\end{CJK} (panel c). Translations, in order: \textit{green}, \textit{olive}, \textit{orange}, \textit{soil}, \textit{bright}, \textit{pale}.}
\label{fig:colors_Chinese} 
\end{figure*}

\end{document}